\documentclass[sigconf]{acmart}
\AtBeginDocument{%
  \providecommand\BibTeX{{%
    \normalfont B\kern-0.5em{\scshape i\kern-0.25em b}\kern-0.8em\TeX}}}

\usepackage[utf8]{inputenc}
\usepackage{array}
\usepackage{makecell}

\usepackage{subfigure} 
 \usepackage{physics}
 \usepackage{array}

 \copyrightyear{2020}
\usepackage[ruled,vlined]{algorithm2e}
\usepackage{amsmath}
\usepackage{mathtools}
\begin{document}

\title{Understanding Brain Dynamics for Color Perception using Wearable EEG headband}

%\title{Learning Query Workload Resources for Database \\ Optimization Decisions}
%\title{Workload Resource Forecasting for Database Optimization Decisions}
%\title{Database Workload-based Resource Prediction for Elastic Database Management Systems}

%\title{Workload-based resource forecasting for self-driving database management systems}
%\title{Memory Estimation - A Learning over Distribution Approach}


\author {Mahima Chaudhary}
\affiliation{%
\institution{Lassonde School of Engineering}
  \country{York University, Toronto, Canada}
}
\email{cmahima@yorku.ca}
%\author{Marin Litiou}
% \affiliation{%

\author{ Sumona Mukhopadhyay}
\affiliation{%
\institution{Lassonde School of Engineering}
  \country{York University, Toronto, Canada}
}
\email{mukhopas@yorku.ca}

\author{ Marin Litoiu}
\affiliation{%
\institution{Lassonde School of Engineering}
  \country{York University, Toronto, Canada}
}
\email{mlitoiu@yorku.ca}

\author{Lauren E Sergio}
\affiliation{
  \institution{Faculty of Health}
   \country{York University, Toronto, Canada}
}
\email{lsergio@yorku.ca}
\author{Meaghan S Adams}
\affiliation{
  \institution{Faculty of Health}
   \country{York University, Toronto, Canada}
}
\email{msadams@yorku.ca}

\renewcommand{\shortauthors}{C, Mahima, et al.}

\begin{abstract}
The perception of color is an important cognitive feature of the human brain. The variety of colors that impinge upon the human eye can trigger changes in brain activity which can be captured using electroencephalography (EEG). In this work, we have designed a multiclass classification model to detect the primary colors from the features of raw EEG signals. In contrast to previous research, our method employs spectral power features, statistical features as well as correlation features from the signal band power obtained from continuous Morlet wavelet transform instead of raw EEG, for the classification task. We have applied dimensionality reduction techniques such as Forward Feature Selection and Stacked Autoencoders to reduce the dimension of data eventually increasing the model's efficiency. Our proposed methodology using Forward Selection and Random Forest Classifier gave the best overall accuracy of 80.6\% for intra-subject classification. Our approach shows promise in developing techniques for cognitive tasks using color cues such as controlling Internet of Thing (IoT) devices by looking at primary colors for individuals with restricted motor abilities.
\end{abstract}



\keywords{Wearable computing, Machine learning, Brain Computer Interface}


%  \includegraphics[width=\textwidth]{Fig1.png}
%  \caption{Schematic Diagram of the learning problem.}
%  %\Description{.}
%  \label{fig1}

%\settopmatter{printfolios=true}

\maketitle
\section{introduction}
The advancements in sensor technologies have facilitated the growth of wearable headband devices for development in Brain-Computer Interface (BCI) applications. One such device is the Muse 2 headband \footnote{https://choosemuse.com/} which is a portable non-invasive device that allows capturing of EEG signals. In this work, we analyzed the relationship between EEG signals and color stimuli using the Muse 2 headband. The objective was to extract the information (features) from EEG signals to classify or distinguish them based on the red(R), green(G), and blue(B) colors that were used to stimulate the cortical activity. The classification result could be potentially used in an integrated IoT environment where it could be used to control appliances \cite{30,21}. One such application could be, where people with restricted motor ability could switch on/off appliances by looking at a particular color. The study that we do is a proof of concept, it can be extended to other colors also.  However, these applications would require input and effort from specialists in other fields too. The work can also be expanded in the healthcare field where it could be used to detect color blindness \cite{28}.\\
Previously, classification tasks like these have been performed \cite{1,2,3} using sophisticated medical-grade EEG devices with multiple sensors but in our work, we used a simple four-electrode/ channel consumer-friendly device to record the raw EEG signals. The use of Muse headband allowed portability to our work and its integration with IoT. Also contrary to previous approaches, in our study, we used features like power, variance in power, various pairwise cross-correlation features and several other statistical features from the signal band power obtained from continuous Morlet wavelet transform for classification task instead of raw EEG signals or event-related potential (ERP) values. The raw EEG data was preprocessed and features that were important to study the effect of color stimuli on EEG were extracted from the data using digital signal processing techniques. \\
We mainly focused on Alpha and Beta frequency bands as these are most likely to be stimulated when a person is alert, attentive, or concentrating and not performing a high cognitive activity. We employed various linear and non-linear Machine Learning (ML) algorithms namely, K Nearest Neighbors, Support Vector Machine (SVM), Logistic Regression, Random Forest, models like Artificial Neural Networks, and boosting approaches like Gradient Boosting, to perform the three-class classification task. We investigated the classification performance of ML algorithms both on a single person's data (intra-subject) as well as on combining the data from different people (inter-subject). We also applied dimensionality reduction techniques like forward feature selection and stacked autoencoders to increase the performance of the architecture. 
The main research questions addressed in this paper are:
\begin{itemize}
    \item Is it possible to distinguish EEG signals from a four-channel wearable headband, produced by RGB color exposure, by training ML models on features that account for statistical, spectral and correlation properties of EEG?
    \item Can feature reduction techniques like Forward-Feature Selection(supervised) and Autoencoders (unsupervised) make the ML algorithms for EEG classification more efficient?
    \item Does the performance of ML algorithms differ for inter-subject and intra-subject classification?
    
\end{itemize}
The rest of the paper is organized as follows. Section 2 contains the related work. In Section 3 we describe our proposed methodology. Section 4 presents our evaluation metrics followed by experimental results in Section 5. Concluding remarks are presented in Section 6.

\section{related work}
In recent years, researchers have used wearable headbands to analyze the response of EEG under different stimuli. K. Johannesen et. al.\cite{5} used SVM to derive useful EEG features in order to predict working memory performance in schizophrenia and healthy adults. The authors in \cite{6} used a regression model trained on data gathered from cognitive tasks (collected from a 6-channel EEG headset ) in order to model mental workload using EEG features for intelligent systems. In \cite{7,8} the EEG data has been used to examine driver's alertness during driving sessions. \\
 Diane Aclo et. al.\cite{9} used a 14 channel EEG device to monitor the effect of color stimuli on people. They used features like power spectral density and waveform length for classification using an Artificial Neural network. In \cite{10} feature selection algorithm has been investigated for EEG signal due to RGB colors using screwable gold EEG electrodes. Arnab Rakshit et. al. \cite{11} proposed the use of a fuzzy space classifier to discriminate colors from EEG by using a 10 electrode device. In \cite{12}, an Emotiv headset has been used to study separation and classification of EEG response to color stimuli by using SVM. Zhang et. el.  in \cite{13} showed how alpha and beta band powers are affected by stimuli from RGB colors.  All the above classification tasks have been conducted using complicated EEG devices in contrast to our work. Furthermore, our proposed method achieves a high accuracy using the Muse headband. Recently, the use of portable headband devices have gained popularity due to their ease of use and accessibility. The authors in \cite{21,40} have used four channel G.tec’s MOBIlab four channel portable device in a problem similar  to ours.  However, their method yielded a lower accuracy of 58\% in comparison to our proposed approach. A headband from Mindwave Neurosky has also been applied \cite{22} in a task similar to our. However, the authors achieved a lower accuracy of 53\% with their method. Our results have shown significant improvement. 
In many studies, Muse has also been used to acquire EEG signals for various classification tasks. EEG-based excitement detection in immersive environments has been studied by Jason et. al in \cite{14}. Krigolson et. al. \cite{15}used Muse headband to Assess Human visual attention by assigning subjects an "oddball"task wherein they saw a series of infrequently  and frequently appearing circles and were instructed to count the number of target circles that they saw. However they did not apply any ML model in their work. In \cite{19} classification task has been performed to classify recreational and instructional video sessions using Muse. They used spectral power and connectivity features from raw EEG in their work and got the best performance with SVM and Logistic Regression model.

\section{proposed methodology}
The main frequencies captured by EEG data are in form of specific human EEG signals namely Delta with frequency 3Hz or below (Deep dreamless sleep), Theta with frequency from 3.5-8 Hz (Deep meditation), Alpha with frequency 8-12 Hz (Calm relaxed yet alert state), Beta with frequency 13-30 Hz (Active, busy thinking) and Gamma with a frequency greater than 41 Hz (Higher mental activity)\cite{29}. Each type of frequency band signal represents a different state of consciousness of mind ranging from sleep to active thinking. We mainly focused on Alpha and Beta frequency bands. The work in this paper has been accomplished in the following five phases: data acquisition, data cleaning and preprocessing, feature extraction, dimensionality reduction, and classification of data into red, green, and blue. We shall explain each component in detail.  
\subsection{Data Acquisition}
Muse headband consists of four channels/electrodes namely AF7 and TP9 on the left and AF8 and TP10 on the right. These are named and positioned according to the International 10-20 System \footnote{https://en.wikipedia.org/wiki/10\%E2\%80\%9320\_system\_(EEG)}, as shown in Figure 1. The sampling rate of Muse is 256Hz. The data from all channels was collected. There were eight subjects (aged 18-30yrs) who participated in the visual experiment. The experiment was conducted using the University of Nottingham's Psychopy 3 \cite{23} toolbox. Five trials each four minutes long were conducted for each participant at different times. In each trial, a color from RGB was shown in a random order, twenty times each, for a period of two seconds each, such that a black color was shown for two seconds between each of the RGB colors to provide a baseline to the experiment. The experiment was conducted in a dark room and the subjects were told to do minimum facial movements and eye blinks. A similar protocol has been followed in previous experiments too \cite{9,30,31}. The time period of the stimulus or the main color was kept small to only capture the effect of color on the cortical excitability. The data from Muse headband was collected using Muse SDK\footnote{https://choosemuse.com/development/} and a third party application for Muse called Mind Monitor\footnote{https://mind-monitor.com/}. The Mind Monitor application indicates potential jaw clenches and eye blinks in the EEG data. We used this capability as a marker to get the starting timestamp of the data. The experiment started with a jaw clench which was captured by Mind Monitor and from that time stamp, the data was separated according to the color stimuli. The architecture we used is shown in Figure 2 and the detailed experimental setup is shown in Figure 3.
\begin{center}
\includegraphics[width=8cm]{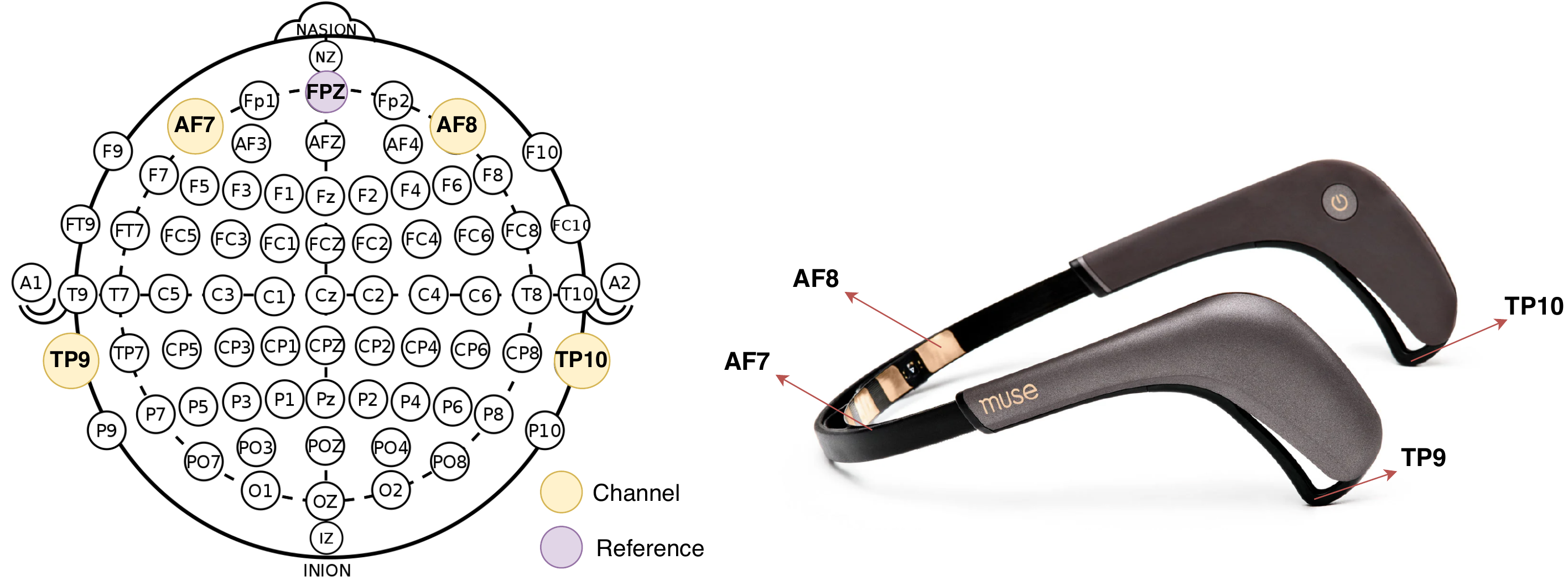}
\captionof{figure}{The 10-20 system of electrode placement for Muse}
\end{center}
\begin{figure*}
  \includegraphics[width=\textwidth,height=6.5cm]{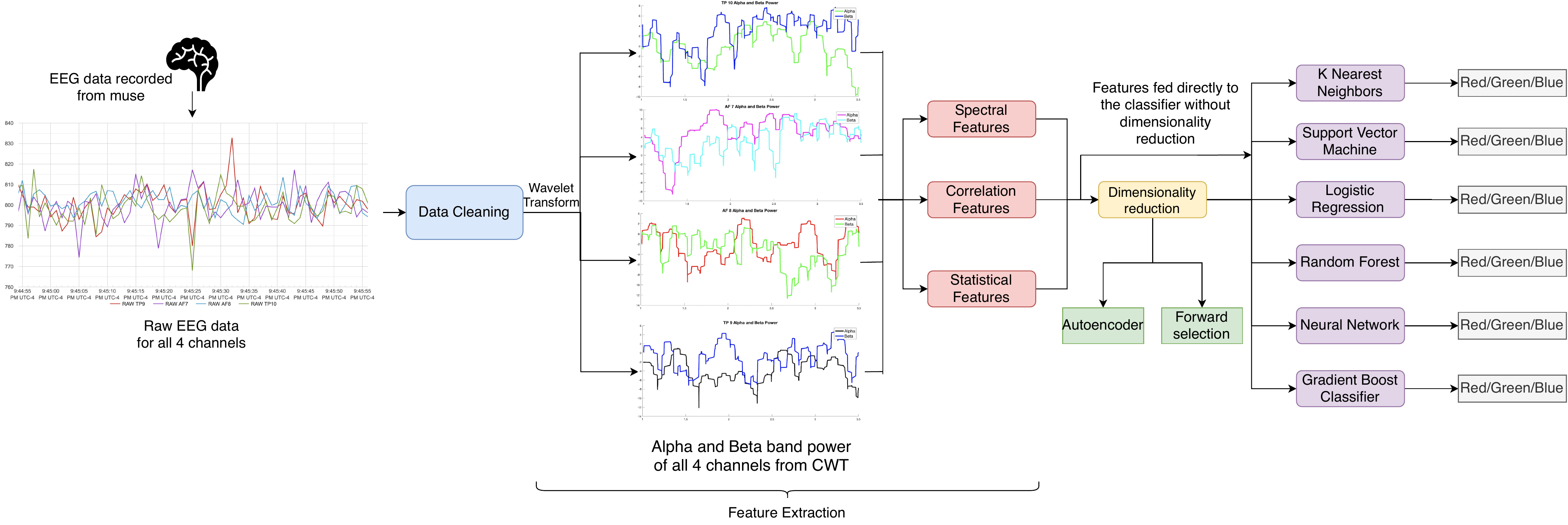}
  \caption{The architecture used in the methodology.}
\end{figure*}

\begin{center}
\includegraphics[width=8.5 cm]{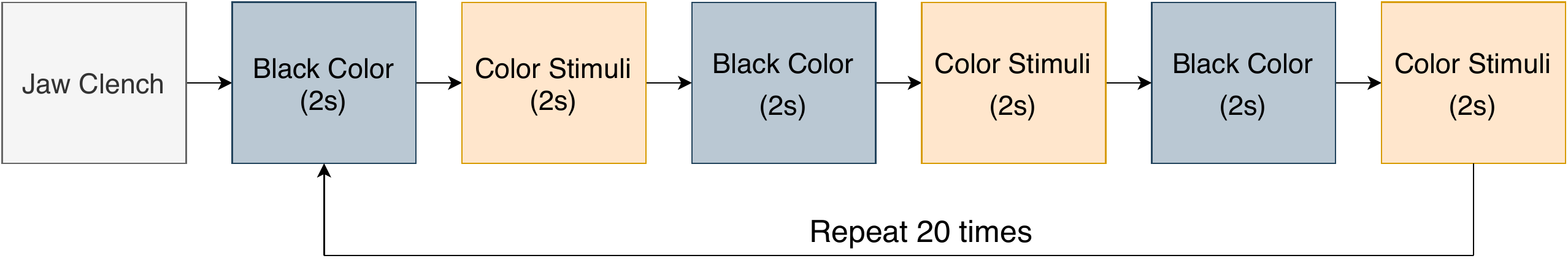}
\captionof{figure}{The experiment protocol for data acquisition}
\end{center}

\subsection{Data cleaning and preprocessing} 
The raw EEG data is generally very noisy and it needs to be cleaned and pre-processed in order to remove artifacts from it. In our methodology, we cleaned the data in two steps. Firstly we analyzed the data using Matlab's EEG lab software \cite{18} and labeled any visible unwanted spikes and noise manually from the data. Secondly, we divided the data into small time windows of 50 ms and computed the variance of data in each window, if it was more than a selected threshold then the time window was flagged. We also examined the individual subject's data and used the trial that has a minimum number of jaw clenches and eye blinks for further experimentation.

\subsection{Feature Extraction}
Feature extraction is a very vital part of our problem. The use of raw EEG data did not give good results in our experiment and so we used Time-Frequency analysis to find frequency band coefficients that were most relevant for our problem i.e. Alpha coefficients(8-12Hz) and Beta coefficients(13-30Hz). In past works, \cite{19,24} Discrete wavelet transform(DWT) has been used to extract the frequency bands of interest. However in our case, we were not interested in all the frequency bands, instead, we only considered alpha and beta bands. The use of DWT would have given us an improper breakdown of bands with the Alpha band in the range of 8-16Hz and beta in range of 16-32Hz and therefore to avoid this we used Continuous wavelet transform method as done in \cite{25,26} to extract the bands of interest. The mother wavelet that we used is the Morlet wavelet. The morlet wavelet has a peak in the center after which it tapers to the edges. The complex Morlet wavelet can be obtained by the convolution of a Gaussian with a sine wave and it is represented by the following equation:
\begin{equation}
  w(t,f)=A*exp(-t^2/2\sigma ^2_t) exp(2\pi ft)
\end{equation}

where $t$ is time, A=$(\sigma_t \sqrt\pi)^{-1/2}$, where $\sigma_t$ is duration of the wavelet and $f$ is the frequency of wavelet. 
%\begin{center}
%\includegraphics[width=7 cm]{complexmorletwavelet.png}
%\captionof{figure}{Complex Morlet Wavelet}
%\end{center}
\begin{figure*}
  \includegraphics[width=\textwidth,height=3.5cm]{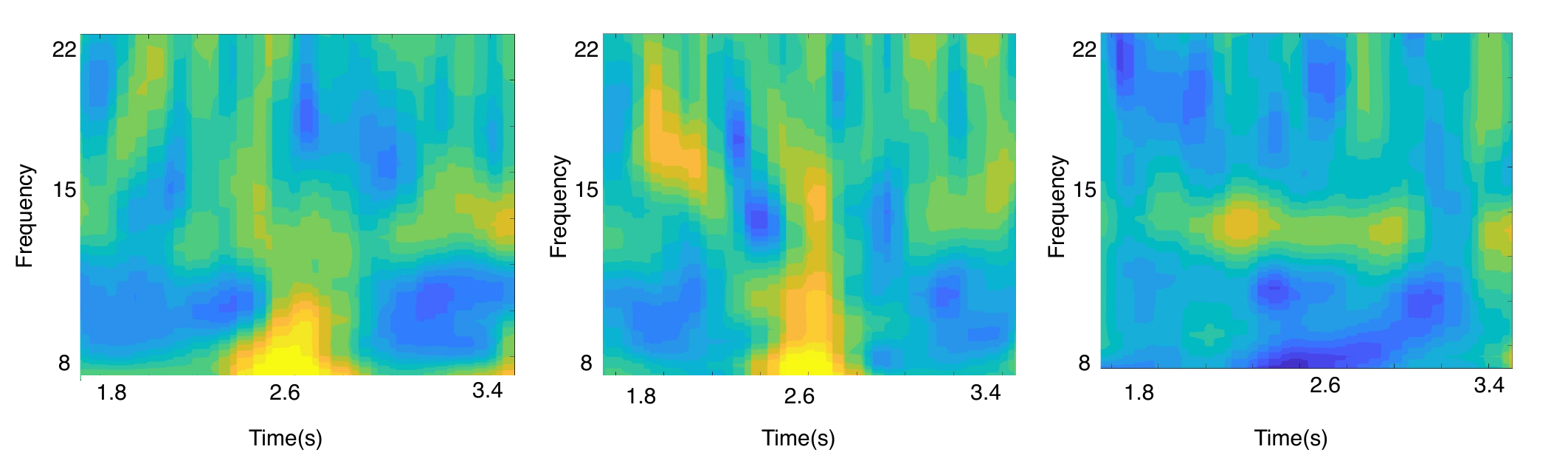}
  \caption{Spectrograms of Red, Green, and Blue respectively. Each spectrogram is made by the EEG obtained from an average of 20 trials (also called Event-related potential) of Subject 1 from channel AF7. The spectrograms show that on the onset of activity at 2.4 sec there is an increase in the power of alpha and low beta frequency band in case of green and red however the power of alpha-band decreases and that of low beta increases in case of blue on the onset of stimuli.}
\end{figure*}
We extracted the power of alpha and beta bands from our EEG signal by convolution of Morlet wavelets of frequencies ranging from 8Hz to 30Hz along the whole signal at each time point. This was done with the help of Fast Fourier Transform (FFT). For each frequency, we first performed FFT of the signal and then the FFT of the Morlet wavelet. We then performed the convolution of the two transformed signals and applied Inverse Fourier transform to get the time-domain representation of data. The magnitude of the complex transformed signal was then extracted and it was squared to obtain the power across all time points. The important thing here is that we rejected the imaginary part as it gave us the phase information and the real part just gave us the band-passed signal but what we were more interested in was the power therefore we extracted the magnitude of the complex signal. We got a spectrogram like representation of the power of the signal, with columns denoting the time points and rows denoting the frequencies from 8Hz to 30Hz. Figure 4 shows the spectrograms for RGB colors. The Morlet wavelet helped to reduce edge artifacts and noise from the data. It also helped to obtain a balance in temporal precision and frequency precision. The sampling rate of the signal and the Morlet wavelet must be the same in order to perform convolution. Figure 5 shows the EEG data with and without the application of the Morlet wavelet. 
We removed the flagged artifacts that we got in Data cleaning step after applying the wavelet transform. This step was done after the transform was applied so that we did not reduce the points below the sampling frequency of 256Hz. The features were extracted from the remaining data. The features accounted for the spectral, correlation as well as statistical properties of the data which were normalized using z-score. We experimented with different time windows of length 100ms, 200ms, 500ms, 1000ms. Each window was taken with a 50\% overlap with the next window. Each window was used to extract a single row of features vector. The next feature vector was obtained by moving the window half of its length.

\begin{center}
\includegraphics[width=5.5cm]{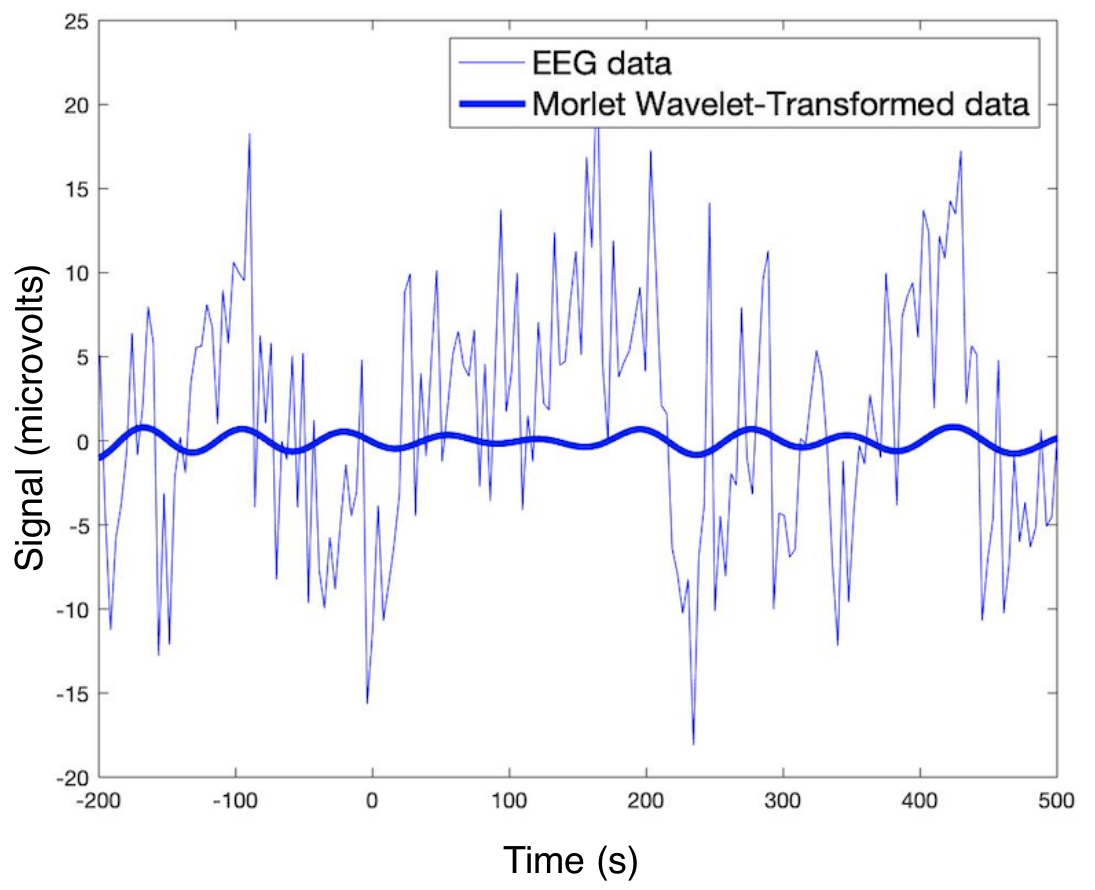}
\captionof{figure}{The original EEG signal and its Morlet-convolution version using a wavelet of 30 Hz}
\end{center}

\subsubsection{Spectral features}
These features were calculated by taking into account the average power of each band, the variance in the power of each band, and the hemispherical difference in each band over a time window for each of the four channels.  Thus we got 18 features(8 average power coefficients for each channel, 8 variance power coefficient for each channel, and 2 hemispheric difference coefficients) for each sample that was formed by a single time window. This method was similar to the one followed in \cite{19}. This set of spectral features are the most commonly used features in many EEG related studies as they allow the model to evaluate any potential changes in the absolute band power due to stimuli.
\subsubsection{Pairwise Correlation features}
In addition to the spectral features, it is also important to study the correlation among different frequency bands from different channels. We calculated this using a pairwise correlation in each time window for each band and each electrode. \cite{19} follows this method too. We got a total of 28 correlation features using this method from a single time window. These features were helpful to find cross-region similarity as some of our data was discontinuous because of artifact removal. 
\subsubsection{Statistical Features}Features that represent the statistical properties of the signal like Kurtosis, Skewness, Shannon Entropy and Hjorth Parameters were also extracted.
\paragraph{Kurtosis, Skewness and Shannon Entropy }
Kurtosis is a measure of outliers in data. Data with less value of Kurtosis has less number of outliers. The Skewness measures the asymmetry in data. The entropy is a measure of information in data. We calculated each of these parameters both for alpha and beta bands, therefore we got 24 features from these properties.
\paragraph{Hjorth Parameters}
They are indicators of statistical properties used in signal processing in the time domain introduced by Bo Hjorth in 1970 \cite{36}. We obtained 16 Hjorth parameters for alpha and beta band for all 4 channels. We calculated two Hjorth parameters namely, the mobility parameter as in equation 2 and complexity parameter as in equation 3 on alpha and beta power bands that we obtain from CWT.
\begin{equation}
 Mobility =\sqrt\frac{var\dv{y(t)}{t}}{var y(t)}
\end{equation}

\begin{equation}
  Complexity =\sqrt\frac{Mobility\dv{y(t)}{t}}{Mobility( y(t))}
\end{equation}

Here $y(t)$ is the alpha or beta band power for a time window. We got a total of 40 statistical features. Table 1 shows all the features obtained from raw EEG data. Figure 6 shows the visualization of features in 2-D space by applying Linear Discriminant Analysis \cite{34}. It shows that the three classes are almost separable.
\begin{center}
\includegraphics[width=7.5 cm]{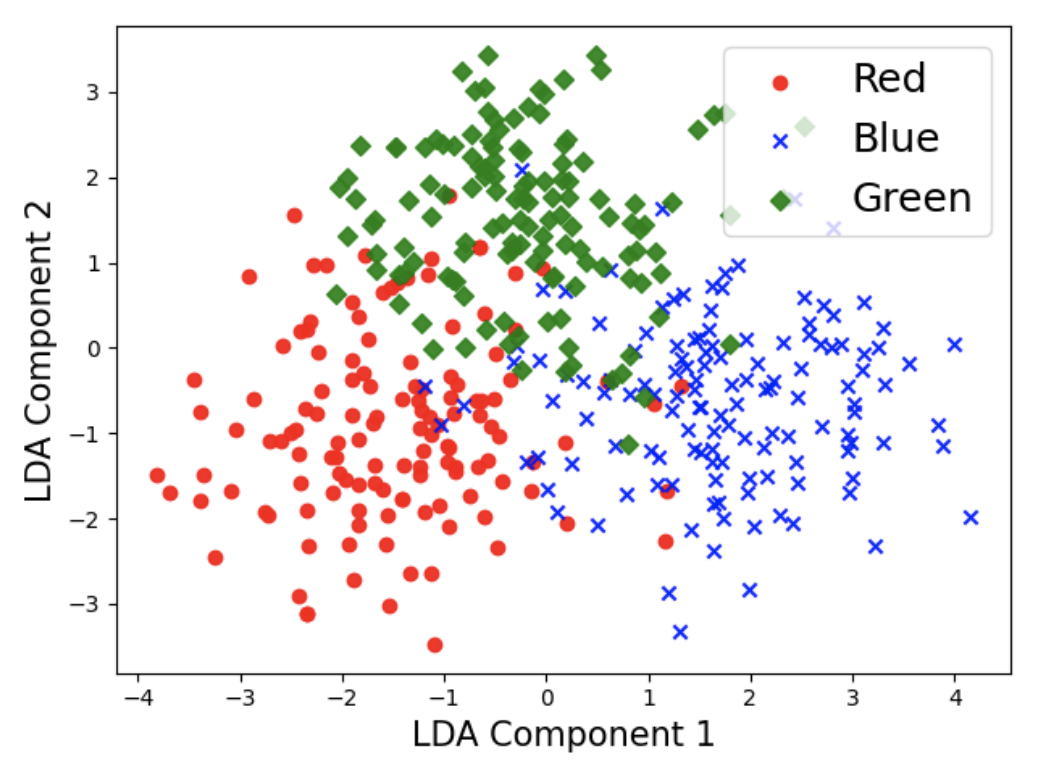}
\captionof{figure}{Visualization of data of Subject 1 for a single trial in 2-D space using Linear Discriminant Ananlysis  }
\end{center}

 %\iffalse
\setlength{\tabcolsep}{0.5pt} % Default value: 6pt
\renewcommand{\arraystretch}{1} 
\begin{center}
  \captionof{table}{The features obtained from raw data.} \label{tab:title}
 \begin{tabular}{||c c c c||} 

 \hline
 \thead{Features \\ \newline} & \thead{Alpha \\\newline} & \thead{Beta\\ \newline} & \thead{Total \\ Features} \\ [0.5ex] 
 \hline
 \makecell{Avg. Power\\features}  & \makecell{TP9, TP10, \\AF7, AF8\\(4)} &\makecell{ TP9, TP10,\\ AF7, AF8\\(4)} & \makecell{8}\\  
 \hline
 \makecell{Var. Power\\features} & \makecell{TP9, TP10, \\AF7, AF8\\(4)} & \makecell{TP9, TP10, \\AF7, AF8\\(4)}& \makecell{8} \\
 \hline
 
\makecell{Hem. diff\\features} & \makecell{|Left hem sensors-\\right hem sensors|\\(1)} & \makecell{|Left hem channels-\\right hem channels|\\(1)} & \makecell{2} \\
 \hline

 \makecell{Correlation\\features} & \makecell{Cross corr of alpha\\ \& beta for all \\channels} & \makecell{Cross corr of alpha\\ \& beta for all \\channels} &\makecell{28} \\
\hline
 \makecell{Kurtosis} & \makecell{TP9, TP10, \\AF7, AF8\\(4)} & \makecell{TP9, TP10, \\AF7, AF8\\(4)}& \makecell{8} \\
 \hline
  \makecell{Sknewness} & \makecell{TP9, TP10, \\AF7, AF8\\(4)} & \makecell{TP9, TP10, \\AF7, AF8\\(4)}& \makecell{8} \\
 \hline
   \makecell{Shannon Entropy} & \makecell{TP9, TP10, \\AF7, AF8\\(4)} & \makecell{TP9, TP10, \\AF7, AF8\\(4)}& \makecell{8} \\
 \hline
   \makecell{Hjorth Parameters} & \makecell{TP9, TP10, \\AF7, AF8\\(8)} & \makecell{TP9, TP10, \\AF7, AF8\\(8)}& \makecell{16} \\
 \hline
\textbf{Total Features} & \makecell{} & \makecell{} & \makecell{\textbf{86}} \\ [1ex] 
  \hline
 
\end{tabular}
 
\end{center}
%\fi

\subsection{Dimensionality reduction}
The process of feature selection is important because it has many advantages like reduced training times, simplified and interpretable models, reduced chances of overfitting i.e. lesser variance and less impact of the curse of dimensionality. We performed feature selection/dimensionality reduction by two different methods. Firstly we used the Forward Feature selection technique which is a supervised approach and secondly, we used Autoencoders \cite{33} which is an unsupervised approach for feature reduction. We elaborate on them in the following subsection.

\subsubsection{Forward Feature Selection}
In this method, we started by selecting one feature and calculating the metric value for each feature on the cross-validation dataset. The feature offering the best metric value was selected and appended to a list of features. The process was repeated next time with two features, one selected from the previous iteration and the other one selected from the set of all features not present in the set of already chosen features. The metric value(f-measure)  was computed for each set of two features and features offering the best metric value were appended to the list of relevant features. This process was repeated until we had the desired number of features. The number of features was reduced to 10 features. The reduced feature set of size ten was chosen after experimenting with feature sets of different sizes, the top ten features gave the best balance between accuracy and number of features.

\subsubsection{Stacked Autoencoders for Feature Extraction}
Autoencoders are neural networks that can be used to reduce the data into a low dimensional latent space by stacking multiple non-linear transformations(layers). They have an encoder-decoder architecture. The encoder maps the input to latent space and the decoder reconstructs the input.
The data in latent space is supposed to have encoded the most important features and has a dimension lesser than the original dimension of data. This data in the latent space can be used as a reduced feature set and the models can be trained on this data. The number of features was reduced to 10(similar to that using forward feature selection) using a stacked autoencoder structure shown in Figure 7. This architecture was chosen after an exhaustive experiment with various architectures.
\begin{center}
\includegraphics[width=2.4cm]{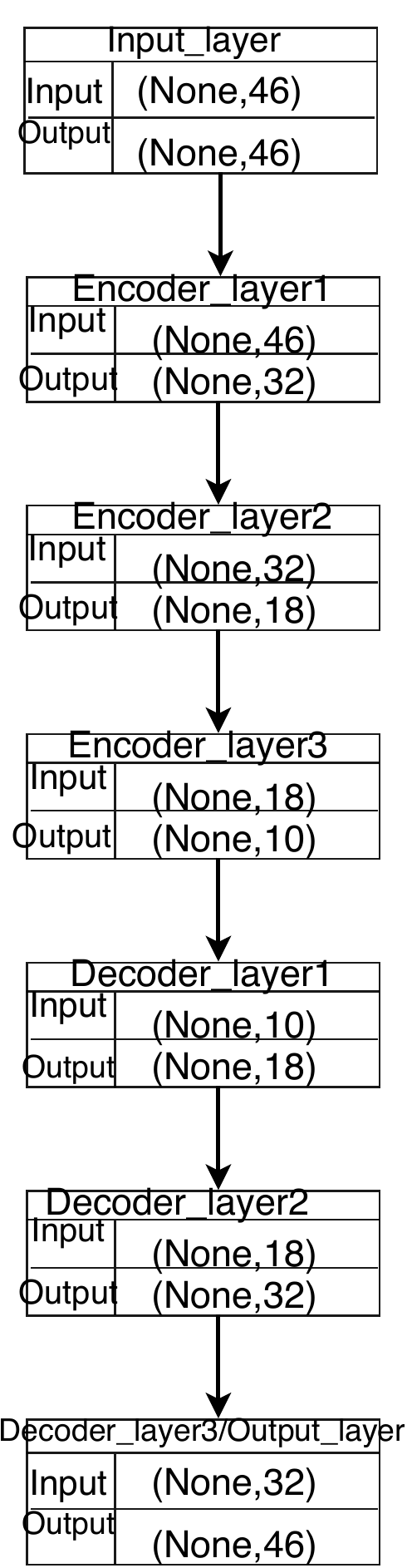}
\captionof{figure}{Final Autoencoder Architecture Used}
\end{center}
\vspace{-2mm}
\subsection{Classification Task}
We applied ML models on the 86 features that we extracted by the procedure explained in Section 3.3. The classification task was done in two folds. We first considered the data from individual subjects and applied models to that data to perform intra-subject classification for which we achieved an accuracy of 80.6\%. Intra-subject classification helped us to study subject-specific differences of the EEG reactivity patterns. Then we considered the combined data from all the subjects and performed inter-subject classification and got an accuracy of 58.1\%. The inter-subject case helped us to make a more generalized model. The classification was done in two ways and their performances have been compared. We performed classification using the original feature set as well as the reduced feature set from forward selection and autoencoders for both intra-subject and inter-subject. Below we elaborate on the models that we have used along with the chosen hyperparameters. We tuned the hyperparameters using Grid Search. The range of hyperparameters chosen was based on previous works \cite{19,21}.

\subsubsection{K Nearest Neighbor (KNN)}
KNN is a non-parametric and lazy learning algorithm. Non-parametric means there is no assumption for underlying data distribution and that's why we tested it in our problem. K is a critical hyperparameter that we varied in the range 4 to 8 in our experiment.  The Euclidean distance was used as the distance metric. As KNN is a lazy learner, therefore it is not advisable to use it in our application, we use it for comparison purposes only.

\subsubsection{Logistic Regression (LR)}
We used logistic regression model both with ridge and lasso regularization. We varied the parameter C or penalty term in the range 0.01 to 100. We found out that lasso regularization gave better results on our data.

\subsubsection{Random Forest (RF)}
The random forest algorithm is an ensemble approach that uses multiple decision trees and makes a classification decision by voting from all the trees. The number of estimators in our problem were varied from 10 to 100.
\subsubsection{Artificial Neural Network (NN)}
We have used ANN with the following architecture: First hidden layer with 300 neurons and second hidden layer with 100 neurons. The activation function used was sigmoid. L2 regularization had been used to avoid overfitting, with a regularization rate of 0.0001. The hyperparameter tuning was done using grid search.

\subsubsection{Support Vector Machine (SVM)}
SVM with RBF kernel has been used in our experiment. The hyperparameters C and Gamma were varied between 0.001 and 100 and 0.01 and 10 respectively.

\subsubsection{Gradient Boosting (GB)}
Gradient boosting is an ensemble learning approach that produces a prediction model in the form of an ensemble of weak prediction models. Gradient boosting combines weak learners into a single strong learner. In our Gradient boosting model we varied the hyperparameter estimators from 10 to 100.

\section{Evaluation Metrics}
Many metrics are used to evaluate ML Models like average accuracy, precision, recall, F-measure, ROC-AUC score, MCC score etc. In our case, we used three metrics for performance evaluation of our models- Average Accuracy, Average ROC-AUC score, and Average Matthews Correlation Coefficient (MCC). 
Since our data is balanced i.e. each class has almost equal representation the average accuracy score would have sufficed but we used the other two additional metrics to verify the performance of our models. We used scikit learn \cite{35} library of python to evaluate the models.
\iffalse
All the three metrics take into account the following parameters: True positives(TP), True Negatives(TN), False Positives(FP) and False Negatives(FN) which can be calculated as follows:
\begin{equation}
    \text{TP}_i=\Sigma \; \textit{\text{samples of class i classified correctly by a classifier} }
\end{equation}
\begin{equation}
    \text{TN}_i=\Sigma \; \textit{\text{samples of other class classified correctly by a classifier} }
\end{equation}
\begin{equation}
    \text{FP}_i=\Sigma \; \textit{\text{samples wrongly classified as class i by a classifier} }
\end{equation}
\begin{equation}
    \text{FN}_i=\Sigma \; \textit{\text{samples wrongly classified as other classes by a classifier} }
\end{equation}
\fi

\subsection{Accuracy Score}
The accuracy score in our problem was calculated as :
\begin{equation}
    \text{Average Accuracy Score} (y,\hat{y})=\frac{1}{n_{sample}}\displaystyle\Sigma_{i=0}^{n_{samples}-1}1(y_i=\hat{y_i} )
\end{equation}

In equation 8,  $\hat{y_i}$ is the predicted value of the i-th sample and $y_i$ is the corresponding true value and $1(x)$ is the indicator function. $n_{samples}$ is the total number of samples.
The accuracy indicates the samples that were correctly classified from all the samples. 

\subsection{ROC-AUC score}
ROC-AUC stands for Receiver operator characteristics- Area under the curve, it basically calculates the area under the receiver operator curve.The ROC curve is created by plotting the true positive rate ($\text{TPR}= \frac{TP}{TP+FN}$) against the false positive rate ($\text{FPR}= \frac{FP}{TN+FP}$) at various threshold settings.
We find the area under the curve to evaluate our model. Since our problem is multiclass therefore we computed the average AUC of all possible pairwise combinations of classes using equation 5 as suggested in \cite{37}.
\begin{equation}
   \text{Average ROC-AUC Score}= \frac{2}{c(c-1)}\Sigma_{j=1}^{c}\Sigma_{k > j}^c (\text{AUC}(j | k) +
\text{AUC}(k | j))
\end{equation}
where $c$ is the number of classes and $\text{AUC}(j | k)$ is the AUC with $j$ as the positive class and $k$ as the negative class and  $\text{AUC}(k | j)$ is vice versa.  In general, $\text{AUC}(j | k) \neq \text{AUC}(k | j))$ in the multiclass case.
\subsection{Matthews Correlation Coefficient}
The Matthews correlation coefficient \cite{27} is used to evaluate the quality of binary and multiclass classifications. The MCC is a kind of correlation coefficient value between -1 and +1. A coefficient of +1 represents a perfect prediction, 0 an average random prediction and -1 an inverse prediction. In the multiclass case, Matthews correlation coefficient can be defined in terms of a confusion matrix $C$ for  $K$ classes. The MCC for multiclass as suggested in \cite{38} is calculated as follows:\\
\begin{equation}
MCC = \frac{
    c \times s - \Sigma_{k}^{K} p_k \times t_k
}{\sqrt{
    (s^2 - \Sigma_{k}^{K} p_k^2) \times
    (s^2 - \Sigma_{k}^{K} t_k^2)
}}
\end{equation}
where $t_k=\Sigma_{i}^{K} C_{ik}$ is the number of times class $k$ truly occurred, $p_k=\Sigma_{i}^{K} C_{ki}$ is the number of times class $k$ predicted, $c=\Sigma_{k}^{K} C_{kk}$ is the total number of samples correctly predicted and $s=\Sigma_{i}^{K} \Sigma_{j}^{K} C_{ij}$ the total number of samples.

\begin{figure*}
  \includegraphics[width=\textwidth,height=3.75cm]{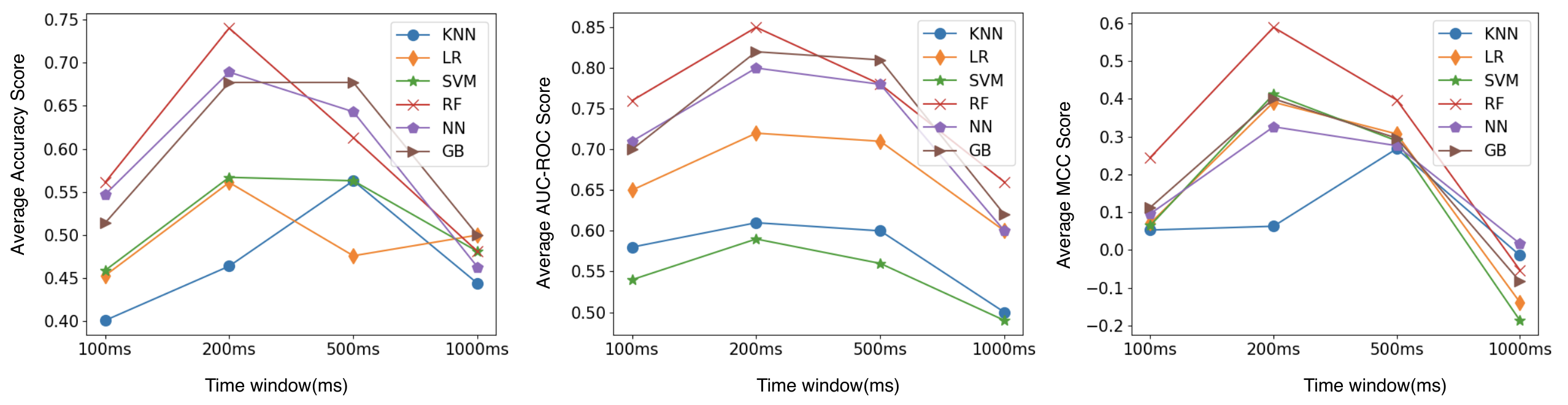}
  \caption{Average Accuracy, Average ROC-AUC score and Average MCC score for different time windows for intra-subject classification}
\end{figure*}
\begin{figure*}
  \includegraphics[width=\textwidth,height=3.75cm]{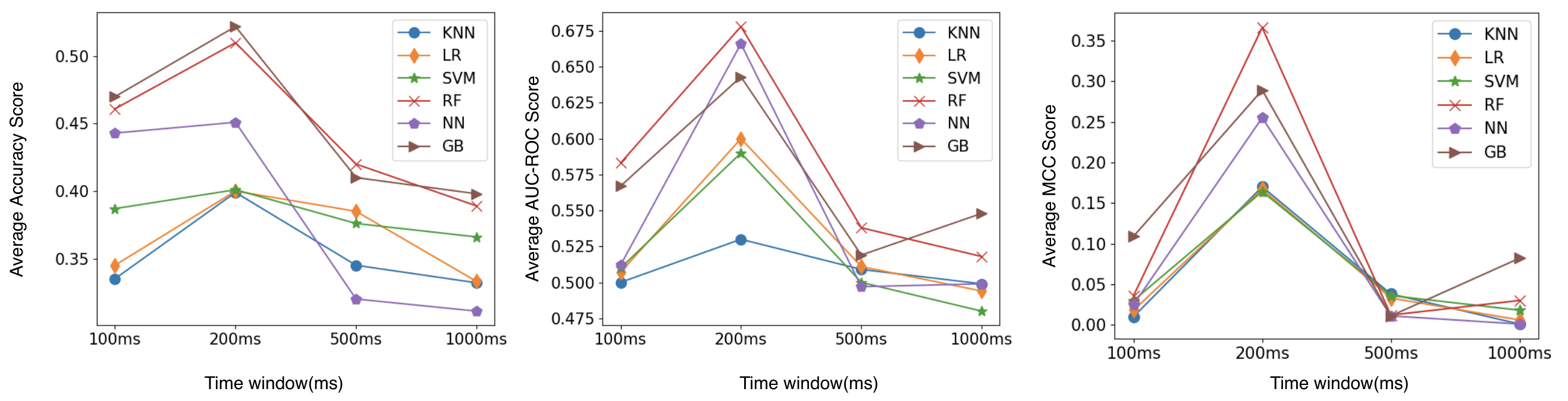}
  \caption{Average Accuracy, Average ROC-AUC score and Average MCC score for different time windows for inter-subject classification}
\end{figure*}

\section{Experiment Results}
We performed two categories of classification namely, intra-subject and inter-subject classification. For intra-subject classification, we applied 5-folds cross-validation for data from five trials of each subject and in the second classification, we applied leave out one subject cross-validation where we trained the model on seven subjects data and validated it using a single subject data and we repeat it for all subjects. The performance metrics that we used to evaluate our model are average cross validation accuracy, average ROC-AUC score, and average MCC. Although average accuracy score should have sufficed our evaluation as our problem is a balanced classification problem i.e. we have almost the same instances for all classes yet we used ROC-AUC and MCC score to verify the results from average accuracy. We report the results on the complete dataset as well on the reduced dataset from dimensionality reduction techniques that we mentioned in Section 3.4. The average accuracy, average AUC score, and average MCC score with different time windows for both intra-subject and inter-subject classification are shown in Figure 8 and Figure 9 respectively. 

We got the best results for a time window of 200ms. We considered the time window of 200ms for further experimentation. We discuss the results in three segments, the results without dimensionality reduction, results after dimensionality reduction from the forward selection algorithm and results after dimensionality reduction from Autoencoder. In the following subsections, we show the results in Table 2-11, considering the average metric score of all the subjects, the best metric score among all subjects and the inter-subject metric score for the three metrics explained in Section 4 . In all the tables the number in brackets is the standard deviation. We have highlighted the highest metrics for each case in all tables. The code for all the experiments is available online \footnote{https://github.com/cmahima/MuseProject}.

\subsection{Results without dimensionality reduction}
 Table 2 shows the accuracy score by using all features. We saw that the Random Forest algorithm performed the best and Neural Network and Gradient Boosting classifier also showed comparable results. The highest accuracy for an individual was 70.2\% which was reasonably better than the accuracy of random guess i.e. 33\%. The inter-subject accuracy of 56.8\% was also very promising considering the fact that we applied leave one subject out cross-validation in this case. The results were better for intra-class classification which means that a customized model could be trained on an individual's data and then it can be used for predictions for a particular subject rather than using data from different people which might also cause privacy issues.

\setlength{\tabcolsep}{0.1pt} % Default value: 6pt
\renewcommand{\arraystretch}{1} 
\begin{center}
  \captionof{table}{Accuracy by using all the features at 200ms time window. } \label{tab:title} 

 \begin{tabular}{||c c c c c c c||} 

 \hline
 \thead{Metrics \\ \newline} & \thead{KNN \\\newline} & \thead{SVM\\ \newline} & \thead{Logistic \\ Regression} & \thead{Random \\Forest} & \thead{Neural\\Network}  & \thead{Gradient\\Boost}\\ [0.5ex] 
 \hline
 \makecell{Avg Subject\\Accuracy}  & \makecell{0.414 \\(0.056)} &\makecell{ 0.474\\(0.018)} & \makecell{0.506\\(0.028)} &\thead{0.625\\(0.018) } & \makecell{0.523\\(0.013)} &\makecell{0.608\\(0.057)}\\ 
 \hline
 \makecell{Best Subject\\Accuracy} & \makecell{0.513\\(0.021)} & \makecell{0.578\\(0.031)} & \makecell{0.600\\(0.022)} & \thead{0.702\\(0.000)}& \thead{ 0.700\\(0.026)} &\thead{0.669\\(0.039)}\\
 \hline
 \makecell{Inter-subject \\Accuracy}& \makecell{0.338\\(0.033)} & \makecell{0.377\\(0.030)} & \makecell{0.408\\(0.030)} &\thead{0.568\\(0.025)} & \makecell{0.490\\(0.033)}&\makecell{0.472\\(0.040)}\\ [1ex] 
  \hline
 
\end{tabular}

\end{center}

In Table 3 we see the ROC-AUC scores. The highest average score of 0.851 was achieved by the Random Forest classifier, the average score of 0.803 was also better than an AUC score of 0.5 in the case of a random classifier. 
\vspace{3mm}
\setlength{\tabcolsep}{0.1pt} % Default value: 6pt
\renewcommand{\arraystretch}{1} 
\begin{center}
  \captionof{table}{ROC-AUC Score by using all the features at 200ms time window} \label{tab:title} 
 \begin{tabular}{||c c c c c c c||} 

 \hline
 \thead{Metrics \\ \newline} & \thead{KNN \\\newline} & \thead{SVM\\ \newline} & \thead{Logistic \\ Regression} & \thead{Random \\Forest} & \thead{Neural\\Network}  & \thead{Gradient\\Boost}\\ [0.5ex] 
 
 \hline
  \makecell{Avg Subject\\Auc score} & \makecell{0.552\\(0.019)} & \makecell{0.514\\ (0.032)} & \makecell{0.676\\(0.027)} & \thead{0.803\\(0.012)}& \makecell{ 0.696\\(0.035)} &\makecell{0.763\\(0.015)}\\
 \hline
 \makecell{Best Subject\\Auc score}  & \makecell{0.600\\(0.043)} &\makecell{0.620\\(0.057)} & \makecell{0.710\\(0.037)} &\thead{0.851\\(0.020)} & \makecell{0.822\\(0.024)} &\makecell{0.810\\(0.012)}\\ 
 \hline
 \makecell{Inter-subject\\Auc score}& \makecell{0.530\\(0.039)} & \makecell{0.570\\ (0.021)} & \makecell{0.601\\(0.046)} &\thead{0.654\\(0.051)} & \makecell{0.611\\(0.035)}&\makecell{0.643\\(0.034)}\\ [1ex] 
  \hline
 
\end{tabular}

\end{center}
\vspace{2mm}
The MCC scores by using all features are in Table 4. A MCC score of 0 means that the classifier is predicting randomly, in our case the highest MCC score was 0.523 which was much higher than a random prediction. 
\vspace{3mm}
\setlength{\tabcolsep}{0.1pt} % Default value: 6pt
\renewcommand{\arraystretch}{1} 
\begin{center}
  \captionof{table}{MCC score by using all the features at 200ms time window} \label{tab:title} 
 \begin{tabular}{||c c c c c c c||} 

 \hline
 \thead{Metrics \\ \newline} & \thead{KNN \\\newline} & \thead{SVM\\ \newline} & \thead{Logistic \\ Regression} & \thead{Random \\Forest} & \thead{Neural\\Network}  & \thead{Gradient\\Boost}\\ [0.5ex] 
 
 \hline
  \makecell{Avg Subject\\MCC} & \makecell{0.120\\(0.050)} & \makecell{0.291\\(0.040)} & \makecell{0.310\\(0.041)} & \thead{ 0.433\\(0.033)}& \makecell{ 0.303\\(0.056)} &\makecell{0.433\\(0.033)}\\
 \hline
 \makecell{Best Subject\\MCC}  & \makecell{0.203\\(0.102)} &\makecell{0.333\\(0.070)} & \makecell{0.318\\(0.041)} &\thead{0.523\\(0.097)} & \makecell{0.431\\(0.070)} &\makecell{0.479\\(0.091)}\\ 
 \hline
 \makecell{Inter-subject\\MCC}& \makecell{0.207\\(0.072)} & \makecell{0.1645\\(0.057)} & \makecell{0.167\\(0.059)} &\thead{0.366\\(0.075)} & \makecell{0.155\\(0.060)}&\makecell{0.189\\(0.058)}\\ [1ex] 
  \hline
 
\end{tabular}

\end{center}
\vspace{2mm}
\subsection{Results with Forward Feature Selection}
We saw significant improvement in the results with the use of forward feature selection which is a supervised feature selection technique. The irrelevant and noisy features were removed and the feature set was reduced to 10. This methodology helped us to curb the overfitting issue too and thus the performance on the validation set improved. In Table 5 we see the average accuracy scores by using the top 10 features. There was an increase in average accuracy by nearly 10\% and we got the highest accuracy of almost 80.6\% which was much better than any other previous approaches that have been used for EEG classification using RGB colors using wearable devices. The average subject accuracy of 72\% showed that the classifier performed well for all the subjects. The average accuracy increased by 9.5\%. In this case, also the results of intra-subject classification were better than that of inter-subject classification. The inter-subject classification accuracy improved by 1.3\%. Random Forest algorithm had given us the best results in this case too with Neural network and Gradient Boost with comparable performance.
\vspace{3mm}
\setlength{\tabcolsep}{0.1pt} % Default value: 6pt
\renewcommand{\arraystretch}{1} 
\begin{center}
 \captionof{table}{Accuracy by using 10 features by forward selection at 200ms time window } \label{tab:title} 
 \begin{tabular}{||c c c c c c c||} 

 \hline
 \thead{Metrics \\ \newline} & \thead{KNN \\\newline} & \thead{SVM\\ \newline} & \thead{Logistic \\ Regression} & \thead{Random \\Forest} & \thead{Neural\\Network}  & \thead{Gradient\\Boost}\\ [0.5ex] 
 \hline
 \makecell{Avg Subject\\Accuracy}  & \makecell{0.492\\(0.038)} &\makecell{0.487\\(0.045)} & \makecell{0.492\\(0.028)} &\thead{0.720\\(0.035)} & \makecell{0.513\\(0.036)} &\makecell{0.597\\(0.048)}\\ 
 \hline
 \makecell{Best Subject\\Accuracy} & \makecell{0.615\\(0.051)} & \makecell{0.604\\(0.028)} & \makecell{0.590\\(0.050)} & \thead{0.806\\(0.041)}& \makecell{ 0.766\\(0.039) } &\makecell{0.720\\(0.035)}\\
 \hline
 \makecell{Inter-subject\\Accuracy}& \makecell{0.377\\(0.013)} & \makecell{0.366\\(0.024)} & \makecell{0.388\\(0.012)} &\thead{0.581\\(0.032)} & \makecell{0.475\\(0.040)}&\makecell{0.411\\(0.019)}\\ [1ex] 
  \hline
 
\end{tabular}
 
\end{center}

We see in Table 6 the AUC scores after forward feature selection. The best AUC score increased by 0.037 and the average AUC score has increased by 0.054. The MCC scores with forward feature selection are in Table 7 which also increased. Thus forward feature selection not only made our architecture efficient computationally but also increased the overall performance of the architecture. In fact, we got the best accuracy of 80.6\% with the use of the Random Forest classifier with forward selection.
\setlength{\tabcolsep}{0.1pt} % Default value: 6pt
\renewcommand{\arraystretch}{1} 
\begin{center}
  \captionof{table}{ROC-AUC Score by using 10 features by forward selection at 200ms time window} \label{tab:title} 
 \begin{tabular}{||c c c c c c c||} 

 \hline
 \thead{Metrics \\ \newline} & \thead{KNN \\\newline} & \thead{SVM\\ \newline} & \thead{Logistic \\ Regression} & \thead{Random \\Forest} & \thead{Neural\\Network}  & \thead{Gradient\\Boost}\\ [0.5ex] 
 
 \hline
  \makecell{Avg Subject\\Auc score} & \makecell{0.546\\(0.017)} & \makecell{0.588\\(0.034)} & \makecell{0.688\\(0.014)} & \thead{0.857\\(0.031)}& \makecell{ 0.699\\(0.033)} &\makecell{0.740\\(0.037)}\\
 \hline
 \makecell{Best Subject\\Auc score}  & \makecell{0.775\\(0.018)} &\makecell{0.670\\(0.018)} & \makecell{0.712\\(0.045)} &\thead{0.901\\(0.013)} & \makecell{0.879\\(0.011)} &\makecell{0.860\\(0.015)}\\ 
 \hline
 \makecell{Inter-subject\\Auc score}& \makecell{0.540\\(0.023)} & \makecell{0.580\\ (0.062)} & \makecell{0.611\\(0.026)} &\thead{0.676\\(0.023)} & \makecell{0.610\\(0.025)}&\makecell{0.632\\(0.014)}\\ [1ex] 
  \hline
 
\end{tabular}

\end{center}

\setlength{\tabcolsep}{0.1pt} % Default value: 6pt
\renewcommand{\arraystretch}{1} 
\begin{center}
  \captionof{table}{MCC score by using 10 features by forward selection at 200ms time window} \label{tab:title} 
 \begin{tabular}{||c c c c c c c||} 

 \hline
 \thead{Metrics \\ \newline} & \thead{KNN \\\newline} & \thead{SVM\\ \newline} & \thead{Logistic \\ Regression} & \thead{Random \\Forest} & \thead{Neural\\Network}  & \thead{Gradient\\Boost}\\ [0.5ex] 
 
 \hline
  \makecell{Avg Subject\\MCC} & \makecell{0.150\\ (0.030)} & \makecell{0.289\\(0.072)} & \makecell{0.302\\(0.037)} & \thead{ 0.578\\(0.061)}& \makecell{ 0.437\\(0.031)} &\makecell{0.310\\(0.061)}\\
 \hline
 \makecell{Best Subject\\MCC}  & \makecell{0.531\\(0.054)} &\makecell{0.346\\(0.054)} & \makecell{0.364\\(0.067)} &\thead{0.638\\(0.013)} & \makecell{0.553\\(0.068)} &\makecell{0.515\\(0.056)}\\ 
 \hline
 \makecell{Inter-subject\\MCC}& \makecell{0.017\\(0.112)} & \makecell{0.189\\(0.051)} & \makecell{0.197\\(0.017)} &\thead{0.476\\(0.015)} & \makecell{0.255\\(0.012)}&\makecell{0.289\\(0.043)}\\ [1ex] 
  \hline
 
\end{tabular}

\end{center}

\subsection{Results with Autoencoder}
We applied autoencoder to observe how an unsupervised feature reduction technique would work on our data. With the autoencoder, a reduced feature set of 10 was obtained. Using this reduced feature set as input to the ML models, we achieved a lower average CV accuracy in comparison to classification using forward feature selection. Therefore autoencoders are not recommended for our application. Table 8-10 show the metrics achieved with the use of autoencoders.
\vspace{1mm}
\setlength{\tabcolsep}{0.1pt} % Default value: 6pt
\renewcommand{\arraystretch}{1} 
\begin{center}
  \captionof{table}{Accuracy by using 10 features by Autoencoder at 200ms time window} \label{tab:title} 
 \begin{tabular}{||c c c c c c c||} 

 \hline
 \thead{Metrics \\ \newline} & \thead{KNN \\\newline} & \thead{SVM\\ \newline} & \thead{Logistic \\ Regression} & \thead{Random \\Forest} & \thead{Neural\\Network}  & \thead{Gradient\\Boost}\\ [0.5ex] 
 
 \hline
  \makecell{Avg Subject\\Accuracy} & \makecell{0.398\\(0.010)} & \makecell{0.362\\(0.026)} & \makecell{0.393\\(0.022)} & \thead{0.430\\(0.011)}& \makecell{ 0.409\\(0.009)} &\makecell{0.417\\(0.000)}\\
 \hline
 \makecell{Best Subject\\Accuracy}  & \makecell{0.417\\(0.000))} &\makecell{ 0.406\\(0.000)} & \makecell{0.434\\(0.000)} &\makecell{0.489\\(0.008) } & \makecell{0.473\\(0.024)} &\thead{0.510\\(0.000)}\\ 
 \hline
 \makecell{Inter-subject\\Accuracy}& \makecell{0.358\\ (0.012)} & \makecell{0.348\\(0.027)} & \makecell{0.347\\(0.023)} &\makecell{0.397\\(0.017)} & \makecell{0.355\\(0.028)}&\thead{0.398\\(0.012)}\\ [1ex] 
  \hline
 
\end{tabular}

\end{center}
\vspace{1mm}
\setlength{\tabcolsep}{0.1pt} % Default value: 6pt
\renewcommand{\arraystretch}{1} 
\begin{center}
  \captionof{table}{ROC-AUC Score by using 10 features by Autoencoder at 200ms time window} \label{tab:title} 
 \begin{tabular}{||c c c c c c c||} 

 \hline
 \thead{Metrics \\ \newline} & \thead{KNN \\\newline} & \thead{SVM\\ \newline} & \thead{Logistic \\ Regression} & \thead{Random \\Forest} & \thead{Neural\\Network}  & \thead{Gradient\\Boost}\\ [0.5ex] 
 
 \hline
  \makecell{Avg Subject\\Auc score} & \makecell{0.501\\(0.037)} & \makecell{0.499\\(0.012)} & \makecell{0.518\\(0.044)} & \thead{0.637\\(0.024)}& \makecell{ 0.598\\(0.026)} &\makecell{0.600\\(0.017)}\\
 \hline
 \makecell{Best Subject\\Auc score}  & \makecell{0.655\\(0.000)} &\makecell{0.560\\(0.018)} & \makecell{0.602\\(0.000)} &\makecell{ 0.686\\(0.007)} & \makecell{0.688\\(0.020)} &\thead{0.730\\(0.001)}\\ 
 \hline
 \makecell{Inter-subject\\Auc score}& \makecell{0.505\\(0.019)} & \makecell{0.499\\ (0.062)} & \makecell{0.520\\(0.026)} &\thead{0.548\\(0.018)} & \makecell{0.510\\(0.037)}&\makecell{0.511\\(0.036)}\\ [1ex] 
  \hline
 
\end{tabular}

\end{center}

\vspace{1mm}
\setlength{\tabcolsep}{0.1pt} % Default value: 6pt
\renewcommand{\arraystretch}{1} 
\begin{center}
  \captionof{table}{MCC score by using 10 features by Autoencoder at 200ms time window} \label{tab:title} 
 \begin{tabular}{||c c c c c c c||} 

 \hline
 \thead{Metrics \\ \newline} & \thead{KNN \\\newline} & \thead{SVM\\ \newline} & \thead{Logistic \\ Regression} & \thead{Random \\Forest} & \thead{Neural\\Network}  & \thead{Gradient\\Boost}\\ [0.5ex] 
 
 \hline
  \makecell{Avg Subject\\MCC} & \makecell{0.099\\(0.019)} & \makecell{0.189\\(0.072)} & \makecell{0.191\\(0.026)} & \thead{ 0.225\\(0.019)}& \makecell{0.218\\(0.039)} &\makecell{0.199\\(0.067)}\\
 \hline
 \makecell{Best Subject\\MCC}  & \makecell{0.261\\(0.000)} &\makecell{0.267\\(0.000)} & \makecell{0.213\\(0.000)} &\makecell{0.229\\(0.027)} & \thead{0.301\\(0.038)} &\makecell{0.261\\(0.000)}\\ 
 \hline
 \makecell{Inter-subject\\MCC}& \makecell{0.015\\(0.022)} & \makecell{0.024\\(0.053)} & \makecell{0.022\\(0.051)} &\thead{0.212\\(0.031)} & \makecell{0.120\\ (0.054)}&\makecell{0.115\\(0.022)}\\ [1ex] 
  \hline
 
\end{tabular}

\end{center}
\vspace{2mm}
The final proposed model for our application is that of Random Forest classifier with forward feature selection. In Table 11 we compare our results with previous efforts that have been done to classify EEG signals on basis of color stimuli using wearable EEG devices. In Figure 10 we show the ROC curve for the proposed model with AUC-ROC score of individual classes for all the subjects where 0 represents red, 1 represents green and 2 represents blue.

%\begin{center}
%\includegraphics[width=8 cm]{ROC.png}
%\captionof{figure}{The ROC-AUC curve for our proposed using architecture using %the subject's data with highest accuracy}
%\end{center}

\begin{figure*}
  \includegraphics[width=\textwidth,height=6.25cm]{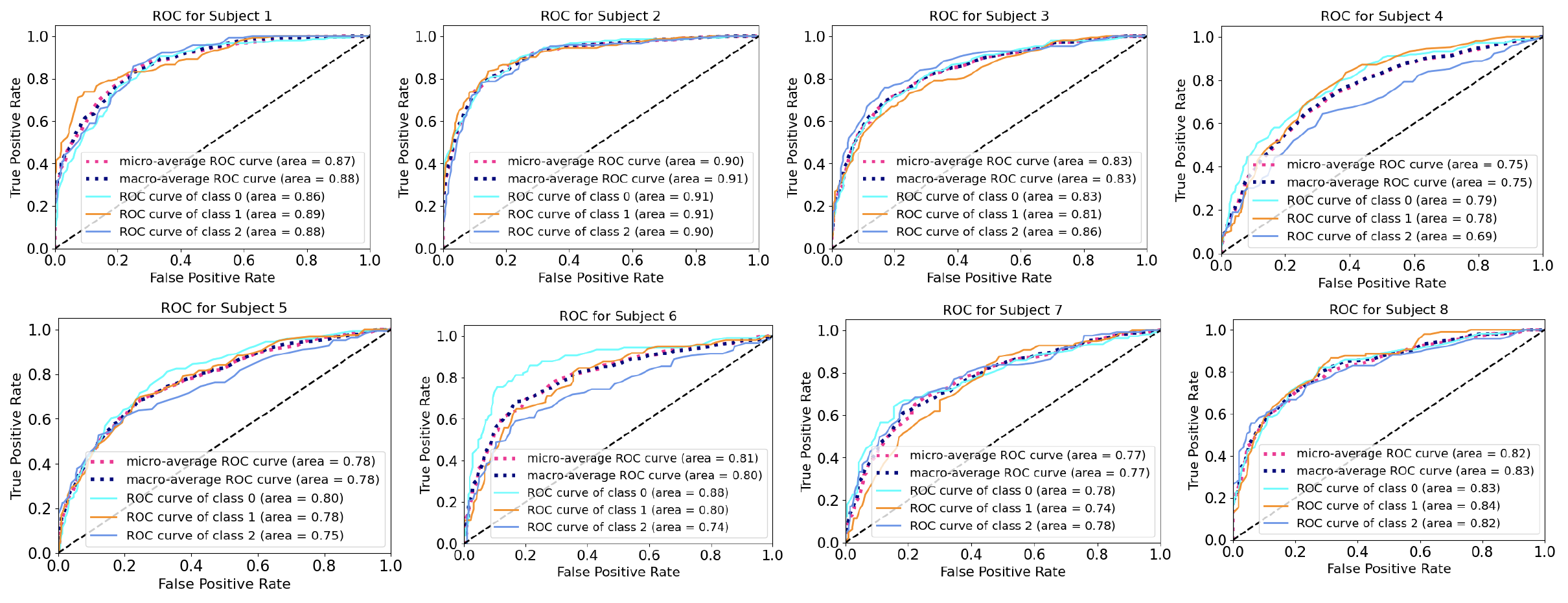}
  \caption{The ROC-AUC curve using our proposed architecture for all the subjects}
\end{figure*}
\vspace{1mm}

\setlength{\tabcolsep}{1pt} % Default value: 6pt
\renewcommand{\arraystretch}{1} 
\begin{center}
  \captionof{table}{Performance of other methods on EEG classification into color stimuli. Our aprroach shows 5-folds average accuracy value for intra-class classification} \label{tab:title} 
 \begin{tabular}{||c c||} 

 \hline
 \thead{Algorithms } & \thead{ Best Average \\ Accuracy } \\ [0.5ex]
 \hline
  \makecell{Martin Angelovski et.al.\cite{22}\\ using 2 channel portable EEG} & \makecell{53\%} \\
 \hline
 \makecell{Sara Åsly et. al.\cite{21,40}\\using 4 channel portable EEG}  & \makecell{58\%} \\ 
 \hline
  \makecell{Kyle Phillips et. al.\cite{12}\\using 14 channel emotiv EEG}  & \makecell{79.6\%} \\ 
 \hline
   \makecell{Rakshit et. al.\cite{11}\\using 10 channel medical EEG}  & \makecell{81.2\%} \\ 
 \hline
 \thead{Our approach using 4 channel portable EEG}& \thead{80.6\%}\\ [1ex] 
  \hline
 
\end{tabular}

\end{center}

\section{conclusions and future work}
We have used EEG signals from a wearable consumer-grade EEG headband to classify the raw EEG data into three classes of colors, red, green, and blue. In our approach, we focussed mainly on Alpha and Beta frequencies and discarded all other lower and higher frequencies which otherwise would have added noise to the data. We extracted various spectral, correlation and statistical features from the data and apply ML models to it. Our proposed model of Random Forest with forward feature selection showed significant improvement when compared to previous approaches. Our methodology achieved an improvement of almost 20\% in the average accuracy of classification.\\ Despite having a fewer number of electrodes Muse performed well in the classification task and gave promising results. The intra-class classification accuracy of 80.6\% shows that wearable devices can be used in integrated IoT frameworks where they can be used in various control applications. The IoT pipeline for this application must take into account the data preprocessing and feature extraction in real-time. The time window for our particular application was small to capture the effect of color stimuli only and avoid unnecessary artifacts in data. This time-window might vary for different applications. One drawback of Muse that we encountered during experiments was that it cannot be worn for a long time due to comfort issues and also the connection can become weak sometimes however one can overcome this problem by applying water to the channels. With the advancement in wearable computing, more comfortable devices are now available that would not bother one if used for a longer time like the new Muse S headband. The Muse 2 device is also sensitive to muscle movements but that is not an issue in our application as we are only interested in a small time window of data when a person focuses on a color. Our work has thus highlighted the capability of these wearable devices to detect and classify the EEG signal on the basis of color stimuli and the results are encouraging. This study opens up a new door to integrate these devices in our day to day lives to use brain signals to control various devices. 

\vspace{-4mm}
\bibliographystyle{unsrt}
\bibliography{bibliography}
 
\end{document}